# Semantic-Driven e-Government: Application of Uschold and King Ontology Building Methodology for Semantic Ontology Models Development

Jean Vincent Fonou-Dombeu[1, 2] and Magda Huisman[2]

[1]Department of Software Studies, Vaal University of Technology, South Africa
fonoudombeu@gmail.com
[2]School of Computer, Statistical and Mathematical Sciences, North-West University, South Africa
Magda.Huisman@nwu.ac.za

## ABSTRACT

*Electronic government (e-government) has been one of the most active areas of ontology development during the past six years. In e-government, ontologies are being used to describe and specify e-government services (e-services) because they enable easy composition, matching, mapping and merging of various e-government services. More importantly, they also facilitate the semantic integration and interoperability of e-government services. However, it is still unclear in the current literature how an existing ontology building methodology can be applied to develop semantic ontology models in a government service domain. In this paper the Uschold and King ontology building methodology is applied to develop semantic ontology models in a government service domain. Firstly, the Uschold and King methodology is presented, discussed and applied to build a government domain ontology. Secondly, the domain ontology is evaluated for semantic consistency using its semi-formal representation in Description Logic. Thirdly, an alignment of the domain ontology with the Descriptive Ontology for Linguistic and Cognitive Engineering (DOLCE) upper level ontology is drawn to allow its wider visibility and facilitate its integration with existing metadata standard. Finally, the domain ontology is formally written in Web Ontology Language (OWL) to enable its automatic processing by computers. The study aims to provide direction for the application of existing ontology building methodologies in the Semantic Web development processes of e-government domain specific ontology models; which would enable their repeatability in other e-government projects and strengthen the adoption of semantic technologies in e-government. The research would be of interest to e-government system developers as well as the Semantic Web community, as the framework and techniques employed to develop the semantic ontology models might be repeated in other domains of knowledge to build ontologies.*

## 1. INTRODUCTION

During the past six years, electronic government (e-government) has been one of the most active areas of ontology development [1], [2], [3], [4], [5], [6], [7]. In e-government, ontologies are being used to describe and specify e-government services (e-services), primarily because they facilitate the semantic integration and interoperability of e-services.

Various different aspects of e-government have been modeled by researchers using the ontology paradigm. For example, Sanati and Lu [6] focus on the integration of e-services while the issue of composition, reconfiguration and evaluation of e-services was addressed by the ONTOGOV project [1], [2]. Other solutions for services integration were proposed in Chen et al. [8] and Gugliotta et al. [9]. Chen et al. [8] proposed a framework for services integration based on specific ontologies, whereas, Gugliotta et al. [9] established the mapping of various ontologies to a predefined e-government system reference model, with the purpose of achieving services integration and interoperability for one-stop portals. The issue of services interoperability is also addressed in [3], [7] and [10] with e-government specific ontology models. Another relevant literature by Puustjarvi [11] proposes a process-document ontology model for the business process modeling in e-government. The Reimdoc project [30] uses various ontologies to model the real-estate transactions.

Most of the aforementioned researches did not state whether any existing ontology building methodology was utilized in the development of their e-government domain specific ontology models. Furthermore, none of these studies employed a framework or an algorithm to show the step by step application of an existing ontology building methodology in a real world e-government service domain. Consequently, it is not clear in these previous works how an e-government system developer without any knowledge of ontology and Semantic Web technologies can repeat these proposed semantic e-government domain specific ontology models in other e-government projects, nor how to develop a solution for a complex public administration system. This oversight hinders the adoption of semantic technologies in e-government.

In this research the Uschold and King [12] ontology building methodology is applied to develop a government domain ontology as an improvement of the work in [13]. In [13], a qualitative data collection technique based on interviews and literature review was employed to gather the business requirements of a government service domain and a framework adopted from the Uschold and King [12] ontology building methodology was applied to construct a government domain ontology. The resulting domain ontology was used in [28] to conduct a case study of Semantic Web development of e-government domain ontology. This previous work is significantly improved in this research. In fact, the framework employed in [13] is revisited and emphases are placed on the detailed description and application of the Uschold and King [12] methodology. The domain ontology developed in [13] is further written semi-formally in Description Logic and validated for semantic consistency. The semi-formal ontology is also coded in OWL to enable its computer processing. Finally, the domain ontology is aligned to the DOLCE upper level ontology to allow its wider visibility and facilitate its integration with existing metadata standard.

The study aims to provide direction for the application of existing ontology building methodologies in the Semantic Web development process of e-government domain specific ontology models; which would enable their repeatability in other e-government projects and strengthen the adoption of semantic technologies in e-government. The research would be of interest to e-government system developers as well as the Semantic Web community, because the framework and techniques we employed to develop the semantic ontology models can be repeated in other domains of knowledge to build ontologies.

The rest of the paper is organized as follows. Section 2 presents the existing ontology building methodologies, the framework employed in this study to build the government domain ontology as well as a case study of its application. Section 3 conducts a discussion and a conclusion is drawn in the last section.

## 2. METHODOLOGY

### 2.1 Ontology Building Methodologies

The commonly used definition of ontology is that proposed by Gruber [14] which states that ontology is an explicit specification of a conceptualization. A conceptualization refers to an abstract and simplified view of a domain of knowledge such as medicine, geology, geographic information system, e-government etc., to be represented for a certain purpose. The domain could be explicitly and formally represented using existing concepts, objects, entities and the relationships that exist between them [14]. Ontologies are widely used in various disciplines including computer science, software engineering, databases and artificial intelligence [15], [16]. In these fields, system developers use an ontology to represent knowledge in a machine processable format to enable their semantic processing by computers.

Several methodologies for building ontology have been proposed in the literature. Detailed comparative analyses of these methodologies are provided in [15], [17], [18]. These methodologies vary in the steps and tasks that they propose a practitioner should perform when building an ontology. There is still no standard method for building ontology. The methodology described in this research follows that of Uschold and King [12]. This methodology was chosen for its clarity and the fact that it is technology and platform independent [17]. The benefit of this methodology is that it is more likely to be understood by novice ontology developers and that it promotes a quicker development of the domain ontology. Furthermore, a new ontology can be constructed from scratch or from existing ontologies [15]. Following the recommendation of Uschold and King [12], a mixture of both approaches has been adopted in this study in the sense that on the one hand completely new domain ontology is constructed, and on the other hand, an alignment of the constructed ontology with the DOLCE upper level ontology [19] is provided. The next section presents the framework for building the government domain ontology.

### 2.2 Framework for Building the Ontology

Uschold and King [12] prescribed five stages for ontology development namely: identify the purpose, building the ontology, evaluation and documentation. In this research the first stage (identify the purpose) is split into two stages: define the purpose as well as the scope of the ontology. In the e-government domain it is important to also delimit the scope and coverage of the desired ontology as a government service for which ontology is being developed may be related to other services within the same department or across other departments. In their work, Uschold and King [12] have further divided the second stage (building the ontology) into three sub-stages namely: ontology capture, ontology coding, and integrating existing ontologies. However, no detailed guidelines are provided on how to gather the concepts and how to determine the relationships between the concepts, "only very vague guidelines relying on brainstorming techniques are given" [17]. In this research the concepts are gathered and the relationships of the domain ontology constructed using a qualitative approach with interviews and a literature review. The domain ontology is coded in a semi-formal representation using Description Logic. The semi-formal ontology is further coded formally in OWL to enable its automatic processing by computers. The second stage of the Uschold and King [12] method is completed by aligning the domain ontology to the DOLCE upper level ontology [19]. The third stage of the Uschold and King [12] method (evaluation) is performed by identifying and fixing semantic inconsistency in the domain ontology. The documentation stage of the Uschold and King [12] method is not executed in this paper because it either does not contribute directly in the building of ontology models which is the main purpose of this study nor does it affect the previous stages used to build the ontology models. Further information on the documentation stage can be found in Uschold and King [12].

Fig. 1 presents the framework used in this study to apply the stages of the Uschold and King [12] methodology presented above in a development projects (DPs) monitoring service domain in Sub Saharan Africa (SSA) and the developing countries at large. Some selected DPs in SSA are provided in Table 1.

## 2.3 Case Study

In this section, the motivation for the case study is presented and the framework in Fig. 1 is applied to develop semantic ontology models including informal/conceptual/domain, semi-formal and formal ontologies.

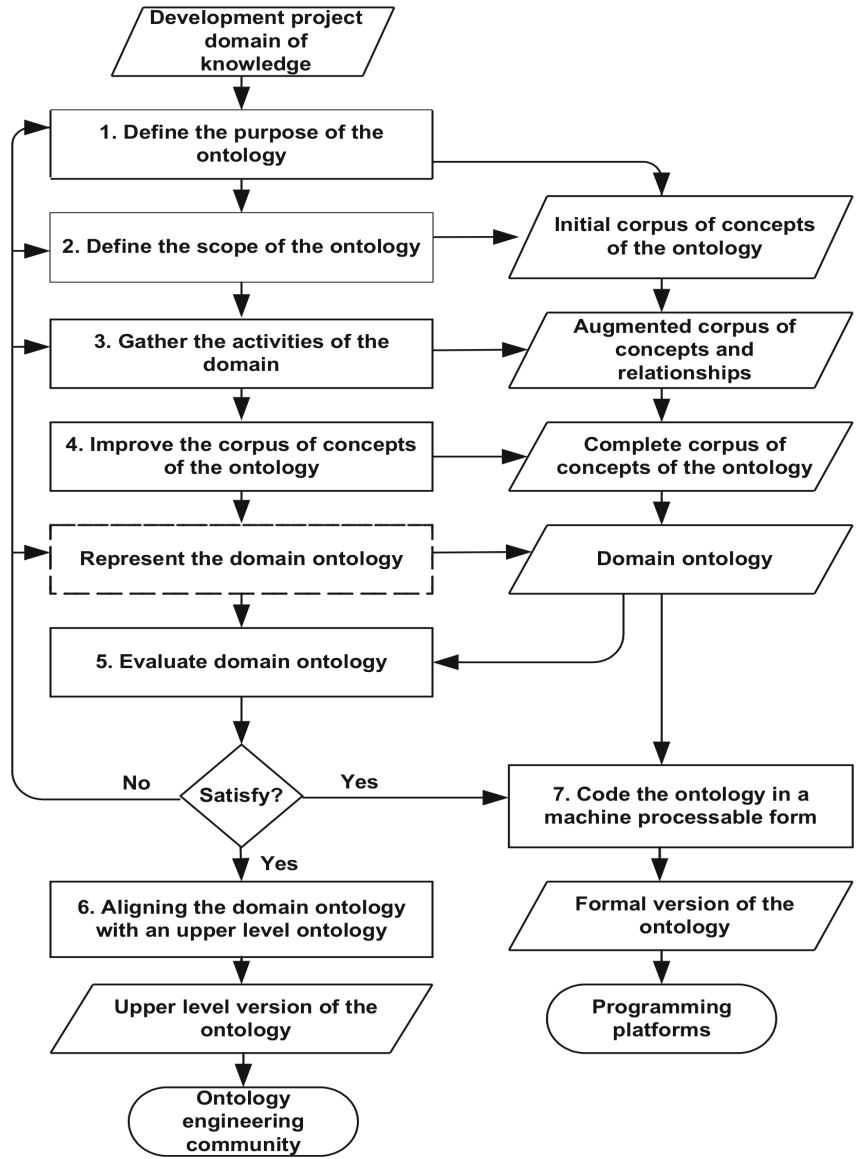

Figure 1: Framework of the ontology building process

### 2.3.1 Motivation

The case study used in this paper is an improved version of the work in Fonou-Dombeu and Huisman [13]. The motivation of the study comes from the fact that, in developing countries and in SSA in particular, almost every government department is somehow involved in the implementation of a programme aiming at improving the welfare of people. These programmes are commonly called development projects (DPs) and include infrastructure development, water supply and sanitation, education, rural development, health care, ICT infrastructure development and so forth. Thus, an e-government web application that could interface all the activities related to DPs implementation in a SSA country would bring tremendous advantages; particularly, such a web application would improve the monitoring and evaluation of DPs and provide transparency, efficiency and better delivery to populations. In Fonou-Dombeu and Huisman [13], an ontology support model for such a web-based e-government application was proposed and used in [28] to conduct a case study of Semantic Web development of e-government domain ontology. This study is a considerable improvement of this previous work. The methodology employed in Fonou-Dombeu and Huisman [13] is revisited and emphases are put on the detailed description and application of the Uschold and King [12] ontology building methodology. Additionally, (1) a set of competency questions are formulated and used to improve the corpus of concepts of the domain ontology, (2) the domain ontology is evaluated for semantic consistency using the Description Logic representation, (3) the domain ontology is further aligned to the DOLCE upper level ontology [19] to allow its wider visibility and facilitate its integration with existing metadata standard [9] and finally, (4) the domain ontology is formally represented in OWL to enable its automatic processing by computers.

As mentioned previously, the framework in Fig. 1 is an extension of the framework used in Fonou-Dombeu and Huisman [13]. Therefore, the first three steps of the framework namely, define the purpose of the ontology, define the scope of the ontology and gather the activities of the domain, are identical and will not be repeated in this research; detailed description of these three first steps are provided in Fonou-Dombeu and Huisman [13]. In summary, in these three first steps, the purposes of the ontology was deduced by analyzing the roles and the current state of impact of DPs in SSA; the scope of the ontology was delimited by analyzing the life cycle of a development project (DP) and the activities that are carried out during various phases. Then, the research was focused on the implementation phase of DPs, which is the phase of the real delivery to people. The activities of the domain were gathered by evaluating case studies of development projects implementation (examples are provided in Table 1), interviewing domain experts including municipalities and non-governmental organizations' (NGO) employees and academic members, and reviewing publications in related fields including project management, project monitoring and evaluation, and capacity building [13]. In light of the above, the following subsections describe the remaining steps of the framework in Fig. 1, from the fourth step till the end.

Table 1: Selected case studies of development projects in SSA

| Case Study No 1 | |
|---|---|
| Project Name | The Women's Information Resource Electronic Service (WIRES) |
| Web Link | *http://www.bridges.org/case_studies/135* |
| Country | Uganda |
| Region | East Africa |
| Financiers | IDRC, World Bank, Hivos |
| Development Sector | Information and Communication technology (ICT) |
| Start Date | 1999 |
| Report Date | 12 November 2003 |
| Status at report date | At the second phase |
| **Implementation** | |
| Approach | Build telecentres to enable women access to ICT and the Internet |
| Key activities | *ICT training, meetings, seminars, workshops* |
| Monitoring and reporting | Formal monitoring every two months, a report after every six months |
| Participants | 90 selected women, Network consultants, local leaders, community members |
| **Case Study No 2** | |
| Project Name | Participatory Design of a Community-Based Child health Information System in South Africa |
| Web Link | *http://www.egov4dev.org/health/case/childhealthis.shtml* |
| Country | South Africa |
| Region | Southern Africa |
| Financiers | *South African Government, UK Department of International Development* |
| Development Sector | Health care |
| Start Date | 2000 |
| Report Date | 2004 |
| Status at report date | *Satisfactory but was still at its infancy* |
| **Implementation** | |
| Approach | Monthly visit of community health workers to households for health data collection |
| Key activities | Household visits, training, interviews, focus group discussions, surveys, evaluation workshops |
| Monitoring and reporting | Regular village health days, use of observation forms during household visits, monthly summary forms for feedback reports |
| Participants | Community workers, traditional leaders, households, community members, municipalities, department of health, stakeholders (childhood practitioners, social workers, etc.) |

## 2.3.2 Improving the Corpus of Concepts of the Ontology

During the first three steps of the framework in Fig. 1 presented above, the initial corpus of concepts of the domain is gathered. This corpus of concepts may not be complete enough to satisfy the purposes of the ontology. Therefore, the actual corpus has to be improved. This can be done by building a set of competency questions which need to be answered by the ontology in order to fulfil its purposes [20]. To this end, a Use Case Diagram (UCD) was designed to represent the interactions of a potential web-based e-government application interfacing the constructed ontology with its target users including project staffs, government authorities, donor organization members, stakeholders and community members. From the UCD, a set of 23 questions is constructed to be answered by the ontology, listed in Table 2. Further, the questions were analysed to find out which concepts are needed in the ontology to enable the inference of appropriate answers to them. This process has added a set of new concepts in the corpus. Finally, the domain ontology of development projects monitoring (OntoDPM) in a SSA country and the developing countries at large is represented in Fig. 2.

It is worth noting that the OntoDPM in Fig. 2 is identical to its first version developed in Fonou-Dombeu and Huisman [13]. In fact, in this previous work, the competency questions were used without any explicit mention of them in the text nor any explanation on how they were constructed. The competency questions in Table 2 are further mapped to the concepts of the formal version of the OntoDPM domain ontology later in this study. In the next section, the domain ontology in Fig. 2 is validated to ensure its semantic consistency.

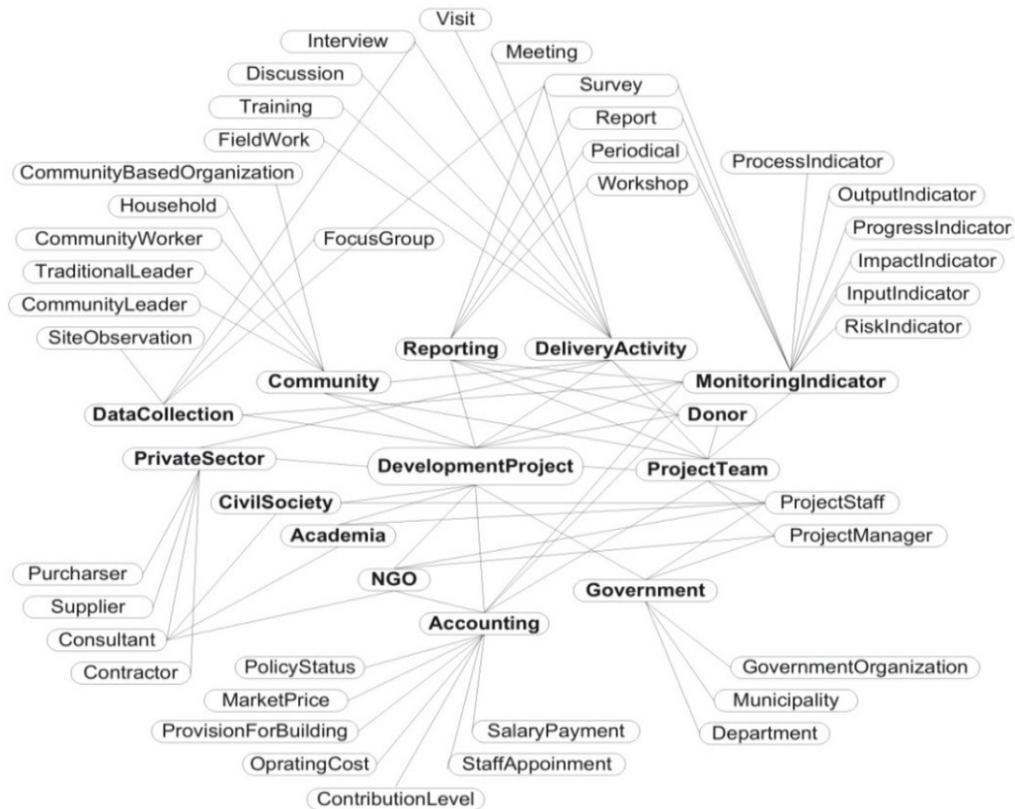

Figure 2: Domain ontology of development projects monitoring in a SSA country and the developing countries at large

### 2.3.3 Evaluate the domain ontology

The ontology engineering field prescribes three layers of ontology development [12]. From a form that can be understood by human beings to one that can be processed by computers, these ontology layers are: informal ontology, semi-formal ontology and formal ontology [12]. The domain ontology, like the one in Fig. 2, is the base ontology model for the development of the formal ontology that can be processed by computers. Therefore, it is important to evaluate its semantic consistency. This can be achieved by creating the semi-formal representation of the domain ontology in Fig. 2 using the Description Logic formalism. Care should then be taken to detect potential semantic inconsistency errors. A semantic inconsistency error is created when a class is wrongly classified as a subclass of another class; or when an instance is wrongly assigned to a concept to which it does not really belong [21]. As drawn in the framework in Fig. 1, semantic inconsistency errors detected at this step can affect the initial graph structure of the domain ontology. Therefore, the ontology specialist should go back and readjust the domain ontology graph to remove the inconsistency errors; this might require a reanalysis of the initial steps applied to build the domain ontology as well (see Fig. 1).

Table 2: List of domain related questions to be answered by the ontology

| Indexes | Questions |
|---|---|
| Q1 | What are the current projects being run in a given locality? |
| Q2 | What is the state of satisfaction of citizens on the implementation of a given DP? |
| Q3 | Who are the stakeholders involved in a given DP? |
| Q4 | Who are the financiers of a given DP? |
| Q5 | Which government division is responsible for the implementation of a given DP? |
| Q6 | What is the list of staff members of a given DP? |
| Q7 | What is the expertise of a DP team member? |
| Q8 | To which institution is a project team member affiliated? |
| Q9 | What is the expertise of a private company involved in a DP? |
| Q10 | What is the amount of contribution from each financier of a DP? |
| Q11 | What is the contribution level of a financier of a DP? |
| Q12 | What is the salary scale of project team members of a given DP? |
| Q13 | What delivery activities are taking place in the day to day run of the DP? |
| Q14 | What are the deadlines for accomplishment of each delivery activity? |
| Q15 | What is the frequency of the delivery activities of DP? |
| Q16 | What are the mechanisms or techniques for data collection during delivery? |
| Q17 | What monitoring and evaluation techniques are used to assess the effectiveness of the implementation of a DP? |
| Q18 | How frequently do the monitoring and evaluation operations take place? |
| Q19 | Who are the actors in charge of the monitoring and evaluation of a DP? |
| Q20 | What reporting techniques are employed to give feedback to top management? |
| Q21 | What are the communication channels between a DP team member and the community? |
| Q22 | Who are the representatives of communities in DPs? |
| Q23 | What are the communication channels between the community and government? |

Two formalisms are commonly used to represent a semi-formal ontology; they include UML class diagram [22] and Description Logic [21], [23]. The OntoDPM domain ontology was written semi-formally in UML in [28]. In this research, we have chosen the Description Logic formalism. The Description Logic representation of domain ontology is useful as it provides strong logical structure for the description and specification of domain knowledge [23], [24], facilitating the detection of semantic inconsistencies in domain ontology [21] and enabling semantic reasoning over the resulting ontology model. Furthermore, the OWL standard which is widely used in the field of Semantic Web is based on Description Logic [24].

Description Logic is a formal language for knowledge representation. Its syntax uses basic mathematical logic symbols such as subset, union, intersection, universal and existential quantifications, etc. to represent the relationships between the constituents of a domain. The Description Logic version of the OntoDPM domain ontology in Fig. 1 is obtained by analysing its semantic and logical structures, identifying its classes, class hierarchy and class instances, and defining relationships between classes. The relationships include inheritance and association/composition relationships. A relationship is also called property or slot. Thereafter, the mathematical logic symbols mentioned above are used to represent the class hierarchy, relationships between classes (inheritance and properties), constraints on properties, etc. For instance, in the class hierarchy of the OntoDPM (see Fig. 2), community worker, community leader, traditional leader, and project staff are people (person class) involved in the development project implementation. Therefore, community worker, community leader, traditional leader, and project staff are subclasses of the person class, representing an inheritance relationship. This relationship is represented in Description Logic formalism using the subset and existential quantification symbols, and the isA property as follows:

$$ProjectStaff \sqsubseteq \exists \ isA.Person$$
$$CommunityWorker \sqsubseteq \exists \ isA.Person$$
$$CommunityLeader \sqsubseteq \exists \ isA.Person$$
$$TraditionalLeader \sqsubseteq \exists \ isA.Person$$

The isA property represents the inheritance relationship between classes. Similarly, the class hierarchy of the OntoDPM shows that department, agency and municipality are division of government. This relationship between government and its divisions can be represented in Description Logic with a hasDivision property, the subset, existential quantification, and union symbols as follows:

$$Government \sqsubseteq \exists \ hasDivision.(Department \cup Agency \cup Municipality)$$

More information on the Description Logic syntax can be found in [21], [23]. Table 3, Table 4, and Table 5 present parts of the semi-formal representation of the OntoDPM domain ontology in Descriptive Logic. In particular, Table 4 and Table 5 represent the class hierarchy of the OntoDPM from which any semantic inconsistency errors have been removed. An alignment of the OntoDPM with the DOLCE upper level ontology is carried out in the next section as the step 6 of the framework in Fig. 1.

Table 3: Important axioms of concepts in the OntoDPM

| |
|---|
| DevelopementProject $\sqsubseteq$ Programme $\Pi$ $\forall$ focuses Community |
| DevlopmentProject $\sqsubseteq$ $\exists$ involves $\geq 1$ (Person $\sqcup$ Financier $\sqcup$ Stakeholder $\sqcup$ Community) |
| DevelopmentProject $\sqsubseteq$ $\exists$ implements $\geq 1$ DeliveryActivity |
| DevelopmentProject $\sqsubseteq$ monitors $\geq 1$ (MonitoringIndicator $\sqcup$ Reporting $\sqcup$ Accounting) |
| ProjectStaff $\sqsubseteq$ Person $\Pi$ $\exists$ affiliates |
| $= 1$ (Municipality $\vee$ Department $\vee$ Agency $\vee$ NGO $\vee$ AcademicInstitution) |
| CommunityWorker $\sqsubseteq$ Person $\Pi$ $\exists$ affiliates Municipality |
| CommunityLeader $\sqsubseteq$ Person $\Pi$ $\exists$ resides Community |
| TraditionalLeader $\sqsubseteq$ Person $\Pi$ $\exists$ resides Community |
| PrivateCompany $\sqsubseteq$ $\exists$ delivers $\geq 1$ DeliveryActivity |
| CommunityBasedOrganization $\sqsubseteq$ owns Community |
| Donor $\sqsubseteq$ Financier $\Pi$ $\forall$ hasContribution ContributionLevel |

Table 4: Axioms of class hierarchy in the OntoDPM

DevelopmentProject ⊑ ∃ hasFunder. Financier
DevelopmentProject ⊑ ∃ hasCorporate. Stakeholder
DevelopmentProject ⊑ ∃ hasPeople. Person
DevelopmentProject ⊑ ∃ hasIndicator. MonitoringIndicator
DevelopmentProject ⊑ ∃ hasActivity. DeliveryActivity
DevelopmentProject ⊑ ∃ hasReport. ReportingTechnique
DevelopmentProject ⊑ ∃ hasData. DataCollectionTechnique
ProjectStaff ⊑ ∃ isA. Person
CommunityWorker ⊑ ∃ isA. Person
CommunityLeader ⊑ ∃ isA. Person
TraditionalLeader ⊑ ∃ isA. Person
Government ⊑ ∃ isA. Financier
Donor ⊑ ∃ isA. Financier
NGO ⊑ ∃ isA. Financier
PrivateCompany ⊑ ∃ isA. Stakeholder
AcademicInstitution ⊑ ∃ isA. Stakeholder
CommunityBaseOrganization ⊑ ∃ isA. Stakeholder
CivilSociety ⊑ ∃ isA. Stakeholder
Consultant ⊑ ∃ isA. PrivateCompany
Contractor ⊑ ∃ isA. PrivateCompany
Supplier ⊑ ∃ isA. PrivateCompany
Purcharser ⊑ ∃ isA. PrivateCompany
Government ⊑ ∃ hasDivision. (Department ⊔ Agency ⊔ Municipality)

Table 5: Axioms of class instances in the OntoDPM

MonitoringIndicator
⊑ ∃ IsIndividualOf. (InputIndicator ⊔ OutputIndicator ⊔ ImpactIndicator ⊔ ProcessIndicator ⊔ ProgressIndicator ⊔ RiskIndicator)
DeliveryActivity ⊑
∃ IsIndividualOf. (Training ⊔ Discussion ⊔ FieldWork ⊔ HouseholdVisit ⊔ Meeting ⊔ Interview)
Reporting ⊑ ∃ IsIndividualOf. (Workshop ⊔ WrittenReport ⊔ Periodical)
DataCollection ⊑ ∃ IsIndividualOf. (Survey ⊔ SiteObservation ⊔ FoucusGroup)

### 2.3.4 Alignment of the Domain Ontology with an Upper Level Ontology

A newly constructed domain ontology should not be kept in isolation; its concepts must further be aligned to those of generic upper level ontologies provided in the ontology engineering field. More information on these upper level ontologies could be found in [19], [25], [29]. This alignment will allow a wider visibility of the constructed domain ontology in the ontology engineering community and facilitate its integration with existing metadata standard [9]. In this research, the concepts of the OntoDPM have been aligned with the DOLCE upper level ontology [19]. Fig. 3 depicts the result of the alignment. The formal version of the OntoDPM is built in the next section and represents the last step (step 7) of the framework in Fig.1.

## 2.3.5 Code the Ontology in a Machine Processable Form

Semantic e-government entails using semantic ontology models to represent and describe government services in such a way that they can be automatically processed by computers. Therefore, the semi-formal version of the domain ontology (see examples in Table 3, Table 4 and Table 5) must be rewritten formally using an existing ontology representation language. The Semantic Web domain provides various languages for the formal representation of ontologies including Extensible Markup Language (XML), Resource Description Framework (RDF), DARPA Agent Markup Language (DAML), and OWL [26]. OWL is the most widely used of these languages because of its high expressive power and the fact that it is the W3C standard ontology language for the Semantic Web [27]. Several software tools are also used for ontology edition including WebODE, OntoEdit, KAON1, and Protégé [15]. Ontology developers prefer Protégé for its ease of use and its abstraction capabilities; it has a graphical user interface which enables ontology developers to concentrate on conceptual modeling without any knowledge of the syntax of the output language [27]. Furthermore, Protégé is open-source software which is downloadable from the Stanford Medical Informatics website.

| | |
|---|---|
| **Dolce:Entity** | **Dolce:Society** |
| **Dolce:Endurant** | Community |
| **Dolce:Physical Endurant** | Civil Society |
| **Dolce:Physical Object** | Community-Based Organization |
| **Dolce:Agentive Physical Object** | **Dolce:Non-Agentive Social Object** |
| Person | Accounting |
| Project Staff | **Dolce:Perdurant** |
| Community Leader | **Dolce:Stative** |
| Community Worker | **Dolce:Process** |
| Traditional Leader | Staff Appointment |
| **Dolce:Non-Physical Endurant** | Salary Payment |
| **Dolce:Non-Physical Object** | Delivery |
| **Dolce:Social Object** | Reporting |
| **Dolce:Agentive Social Object** | Data Collection |
| **Dolce:Social Agent** | **Dolce:Quality** |
| Private Company | **Dolce:Temporal Quality** |
| Contractor | Monitoring Indicator |
| Supplier | Input Indicator |
| Purchaser | Output Indicator |
| Consultant | Process Indicator |
| Government | Progress Indicator |
| Department | Risk Indicator |
| Municipality | Impact Indicator |
| Agency | **Dolce:Abstract Quality** |
| Non-Governmental Organization | Operating Cost |
| Academia | Provision for Building |
| Donor Organization | Contribution Level |

Figure 3: Alignment of the OntoDPM with the DOLCE upper level ontology

In Fonou-Dombeu and Huisman [28], Protégé was used to create the OWL formal version of the OntoDPM domain ontology using its semi-formal representation in UML. In this research, the formal version of the OntoDPM in OWL is created with Protégé using the semi-formal specifications of its class hierarchy, relationships between classes and class instances written in Description Logic and validated for semantic consistency (see Table 3, Table 4 and Table 5). Several concepts including class, individual, slot, domain, range, etc. are employed in Protégé to create ontology. A slot is also called property and represents a relationship between classes. Each slot has a domain and a range, which are the classes involved in the relationship.

Once the ontology has been created with Protégé and saved as an OWL file onto the disc, it appears in the OWL syntax. The OWL syntax provides facilities for representing ontology elements such as inheritance, instance, slots, domain and range of a slot, etc. For instance in the OntoDPM, community worker, community leader, traditional leader, and project staff are subclasses of the person class (see Description Logic example in Table 4). This inheritance relationship is represented in OWL syntax with the following OWL code generated with Protégé.

```
<owl:Class rdf:about="#ProjectStaff">
    <rdfs:subClassOf rdf:resource="#Person"/>
</owl:Class>
<owl:Class rdf:about="#CommunityLeader">
    <rdfs:subClassOf rdf:resource="#Person"/>
</owl:Class>
<owl:Class rdf:about="#CommunityWorker">
    <rdfs:subClassOf rdf:resource="#Person"/>
</owl:Class>
<owl:Class rdf:about="#TraditionalLeader">
    <rdfs:subClassOf rdf:resource="#Person"/>
</owl:Class>
```

Similarly, in the OntoDPM, department, agency and municipality are divisions of government. The relationship between government and its divisions can be represented in Protégé with the slot hasDivision. Then, the domain of the hasDivision slot will be the government class and its ranges, the department, agency, and municipality classes. The hasDivision slot and its domain and ranges are represented in OWL with the following code generated with Protégé.

```
<owl:ObjectProperty rdf:about="#hasDivision">
    <rdfs:domain rdf:resource="#Government"/>
    <rdfs:range rdf:resource="#Agency"/>
    <rdfs:range rdf:resource="#Department"/>
    <rdfs:range rdf:resource="#Municipality"/>
</owl:ObjectProperty>
```

Parts of the OntoDPM properties, class hierarchy and class instances created with Protégé and imported with the programming editor JCreator are depicted in Fig. 4. In Fig. 4, it is shown each property (hasData, hasDivision, etc.) along with its domain and range(s), whereas, the inheritance relationships between classes are represented with the "subClassOf" property; class instances are indicated with the keyword "Thing".

# 3. DISCUSSION

This section (1) provides a short discussion on the use of OWL ontology in e-government (2) maps the competency questions in Table 2 with concepts of the formal version of the OntoloDPM domain ontology and (3) compares this study with related research.

## 3.1 OWL Ontology in e-Government

As mentioned earlier, OWL is a common language employed for semantic knowledge representation in e-government. In particular, OWL ontologies allow the composition [1], [7], searching, matching, mapping and merging [2], [6] of e-government services and facilitate their integration [1], [2], [5], maintenance [1] and interoperability [3], [4], [6], [7]. Therefore, generating OWL ontology from a government service domain as it is done in this research and in [28] is an important step towards the development of  Semantic Web applications as e-government applications, which have potential to perform semantic inference and reasoning over the OWL ontology and facilitate software components integration and interoperability. However, for the OWL ontology to be useful it has to be deployed in a real world application; this requires its storage and access through programming platforms. Future research will investigate the platforms for deploying OWL ontologies in real world applications including Java API, .NET, ASP, etc., as well as the database storage and query mechanisms of OWL ontologies. The mapping of the competency questions in Table 2 with the concepts of the formal version of the OntoDPM domain ontology is carried out in the next section.

## 3.2 Mapping of Competency Questions with Concepts of the Formal Ontology

In Section 2.3.2 the competency questions in Table 2 were analyzed to improve the corpus of concepts of the OntoDPM domain ontology in Fig. 2.  Afterward, the concepts of the domain ontology were further analyzed to build classes, class instances, class hierarchy and properties between classes, for the formal representation (machine processable form) of the domain ontology; this process has added new concepts (e.g. person class) in the domain ontology as well as discarded some of them (e.g. project manager concept has simply become project staff). With all these changes in the transition from the human readable version of the domain ontology to the machine processable one, it is also important to map the competency questions to the concepts of the formal representation of the domain ontology; to ensure that answers to these questions would still be inferred from the ontology. In this section, a quantitative analysis is performed to establish the mapping of the competency questions in Table 2 to concepts of the formal version of the OntoDPM domain ontology. The formal version of the OntoDPM domain ontology is referred to from here on as formal ontology.

**Part of OWL Representation of OntoDPM Properties**

```
<owl:ObjectProperty rdf:about="#hasData">
  <rdfs:range rdf:resource="#DataCollectionTechnique"/>
  <rdfs:domain rdf:resource="#DevelopmentProject"/>
</owl:ObjectProperty>
<owl:ObjectProperty rdf:about="#hasDivision">
  <rdfs:domain rdf:resource="#Government"/>
  <rdfs:range rdf:resource="#Agency"/>
  <rdfs:range rdf:resource="#Department"/>
  <rdfs:range rdf:resource="#Municipality"/>
</owl:ObjectProperty>
<owl:ObjectProperty rdf:about="#hasFunder">
  <rdfs:domain rdf:resource="#DevelopmentProject"/>
  <rdfs:range rdf:resource="#Financier"/>
</owl:ObjectProperty>
<owl:ObjectProperty rdf:about="#hasIndicator">
  <rdfs:domain rdf:resource="#DevelopmentProject"/>
  <rdfs:range rdf:resource="#MonitoringIndicator"/>
</owl:ObjectProperty>
<owl:ObjectProperty rdf:about="#hasPeople">
  <rdfs:domain rdf:resource="#DevelopmentProject"/>
  <rdfs:range rdf:resource="#Person"/>
</owl:ObjectProperty>
<owl:ObjectProperty rdf:about="#hasReport">
  <rdfs:domain rdf:resource="#DevelopmentProject"/>
  <rdfs:range rdf:resource="#ReportingTechnique"/>
</owl:ObjectProperty>
```

**Part of OWL Class Hierarchy of the OntoDPM**

```
<owl:Class rdf:about="#Person">
  <owl:Class rdf:about="#ProjectStaff">
    <rdfs:subClassOf rdf:resource="#Person"/>
  </owl:Class>
  <owl:Class rdf:about="#CommunityLeader">
    <rdfs:subClassOf rdf:resource="#Person"/>
  </owl:Class>
  <owl:Class rdf:about="#CommunityWorker">
    <rdfs:subClassOf rdf:resource="#Person"/>
  </owl:Class>
  <owl:Class rdf:about="#TraditionalLeader">
    <rdfs:subClassOf rdf:resource="#Person"/>
  </owl:Class>
</owl:Class>
<owl:Class rdf:about="#Financier">
  <owl:Class rdf:about="#Donor">
    <rdfs:subClassOf rdf:resource="#Financier"/>
  </owl:Class>
  <owl:Class rdf:about="#Government">
    <rdfs:subClassOf rdf:resource="#Financier"/>
  </owl:Class>
  <owl:Class rdf:about="#NGO">
    <rdfs:subClassOf rdf:resource="#Financier"/>
  </owl:Class>
</owl:Class>
```

**Part of OWL Class Instances of the OntoDPM**

```
<owl:Thing rdf:about="#InputIndicator">
<rdf:type rdf:resource="#MonitoringIndicator"/>
</owl:Thing>
<owl:Thing rdf:about="#OutputIndicator">
<rdf:type rdf:resource="#MonitoringIndicator"/>
</owl:Thing>
<owl:Thing rdf:about="#ProcessIndicator">
<rdf:type rdf:resource="#MonitoringIndicator"/>
</owl:Thing>
<owl:Thing rdf:about="#ProgressIndicator">
<rdf:type rdf:resource="#MonitoringIndicator"/>
</owl:Thing>
<owl:Thing rdf:about="#RiskIndicator">
<rdf:type rdf:resource="#MonitoringIndicator"/>
</owl:Thing>
<owl:Thing rdf:about="#ImpactIndicator">
<rdf:type rdf:resource="#MonitoringIndicator"/>
</owl:Thing>
```

Figure 4: Part of OWL representation of OntoDPM domain ontology

Let $Q$ and $C$ be the sets of competency questions in Table 2 and concepts of the formal ontology respectively. The Semantic Relatedness Score (SRS) of a question Qi, $1 \leq i \leq 23 \subseteq Q$ with concepts of the formal ontology noted $SRS_{Qi}$, is defined as the number of concepts to which Qi is semantically related to in the formal ontology. Mathematically $SRS_{Qi}$ is defined as in equation (1) below.

$$SRS_{Qi} = \sum Card(c_j), 1 \leq j \leq n \qquad (1)$$

Where, $n$ is the number of main concepts to which Qi is related to in the formal ontology and $Card(c_j)$ the number of instances of the concept $c_j \subseteq C$ in the formal ontology. The number $Card(c_j)$ is defined as in equation (2).

$$Card(c_j) = 1 + \sum Card(Instance(c_j)) \qquad (2)$$

Where, $Instance(c_j) \subseteq C$ is an instance of the concept $c_j \subseteq C$ in the formal ontology. Furthermore, $Card(c) = 1$ if the concept $c \subseteq C$ is an instance or a concept which is a leaf in an inheritance hierarchy in the formal ontology.

As an example, let's consider the concept *government* in the domain ontology. The instances of *government* are *department*, *agency* and *municipality*. Based on equations (1) and (2) the SRS of a question $Q \subseteq Q$ which is semantically related to the *government* concept is computed as follows:

$$Card(government) = 1 + Card(department) + Card(agency) + Card(municipality) = 4$$

Similarly, a manual analysis of each competency question in Table 2 was performed and the main concepts of the development project (DP) domain which are semantically related to the question (see column 2 of Table 6) enumerated. For instance, let's consider the first question (Q1) in Table 2: "*What are the current projects being run in a given locality?*" The answer to this question relies on the information that would be provided by the municipality (concept which is semantically related to the word "locality" in the question Q1) as well as the development projects (concept which is semantically related to the word "projects" in the question Q1) under implementation in the municipality. As a result of the above analysis, it appears that the question Q1 in Table 2 is semantically related to the main concepts *development projects* and *municipality* in the development project domain (see column 2 of Table 6); in fact, the concepts *development projects* and *municipality* are synonyms of the words "locality" and "projects" respectively in the question Q1. This analysis was done independently of any knowledge of concepts of the formal ontology. A similar analysis was carried out for all the questions in Table 2 and the result is provided in the column 2 of Table 6.

Thereafter, the equation (1) and equation (2) were applied to compute the SRS of each question (see column 3 of Table 6). For instance, the column 2 of Table 6 shows that the competency question Q1 refers to the concepts *development projects* and *municipality* in the DP domain; both concepts are also concepts of the formal ontology. Furthermore, these concepts do not have any instances in the formal ontology class hierarchy; then, the $SRS_{Q1}$ is 2 (see intersection of row 2 and columns 3 and 4 in Table 6). Similarly, column 2 of Table 6 shows that Q2 refers to concepts *community*, *development projects* and *monitoring indicator* in the DP domain. However, only *development projects* and *monitoring indicator* are also concepts in the formal ontology; in fact, the concept *community* has been discarded when building the semi-formal ontology. Furthermore, the concept *monitoring indicator* has six instances namely: output, input, process, progress, impact, and risk indicators, in the formal ontology (see the bottom part of Fig. 4). Then, the $SRS_{Q2}$ is 8 (see intersection of row 3 and columns 3 and 4 in Table 6). As a result, the chart of the semantic relatedness scores of the competency questions in Table 2 with the concepts of the formal ontology is drawn in Fig. 5; it appears that each competency question is related to at least 2 concepts in the formal ontology. This indicates that the competency questions in Table 2 map the concepts of the formal ontology and that the developed domain ontology would be able to fulfill its purposes as defined at the early stage of its development. The next subsection compares this study with related works.

Table 6: Semantic relatedness score of competency questions with concepts of the formal Ontology

| $Qi$ | $c_i$ | $\sum Card(c_i)$ | $SRS_{Qi}$ |
|------|-------|------------------|------------|
| Q1 | DP, municipality | 1+1 | 2 |
| Q2 | community, DP, monitoring indicator | 0+1+(1+(1x6)) | 8 |
| Q3 | DP, stakeholder | 1+(1+(1x4)+(1x4)) | 10 |
| Q4 | DP, financier | 1+(1+(1x3)+(1x3)) | 8 |
| Q5 | department, municipality, agency, DP | 1+1+1+1 | 4 |
| Q6 | DP, project staff | 1+1 | 2 |
| Q7 | DP, project staff | 1+1 | 2 |
| Q8 | project staff, government, NGO, private sector | 1+(1+(1x3))+1+(1+(1x4)) | 11 |
| Q9 | DP, private sector | 1+(1+(1x4)) | 6 |
| Q10 | financier, DP | 1+(1x3)+(1x3) | 7 |
| Q11 | contribution level, financier, DP | 1+(1+(1x3)+(1x3))+1 | 9 |
| Q12 | salary payment, project staff, DP | 1+1+1 | 3 |
| Q13 | delivery activity, DP | 1+(1x7)+1 | 9 |
| Q14 | delivery activity, monitoring indicator, DP | 1+(1x7)+(1+(1x6))+1 | 16 |
| Q15 | delivery activity, DP, monitoring indicator | 1+(1x7)+1+(1+(1x6)) | 16 |
| Q16 | delivery activity, data collection | 1+(1x7)+(1+(1x4)) | 13 |
| Q17 | monitoring indicator, DP | 1+(1x6)+1 | 8 |
| Q18 | Monitoring indicator, DP | 1+(1x6)+1 | 8 |
| Q19 | DP, monitoring indicator, project staff, private sector, community | 1+(1+(1x6))+1+(1+(1x4)) + 0 | 14 |
| Q20 | reporting technique, DP, | 1+(1x4)+1 | 6 |
| Q21 | DP, project staff, community leader | 1+1+1 | 3 |
| Q22 | community leader, DP | 1+1 | 2 |
| Q23 | community leader, government | 1+(1+(1x3)) | 5 |

## 3.3 Comparison of this Study with Related Research

Table 7 and Fig. 6 draw a comparison of this research with related works. We have focused the comparison on the following criteria:

- Ontology building methodology: We are trying to find out if the research has disclosed the ontology building methodology employed.
- Framework or algorithm: Has the study designed or adopted a framework for the practical application of the ontology building methodology used (if there is any) in a government service domain.
- Level of specification: We seek to find out how detailed the specification of the proposed e-government domain specific ontology model in the study was.
- Implementation details: We are looking at the details of the implementation of the proposed e-government domain specific ontology models in terms of the Semantic web ontology languages and platforms employed, and whether a platform employed is open-source or proprietary.

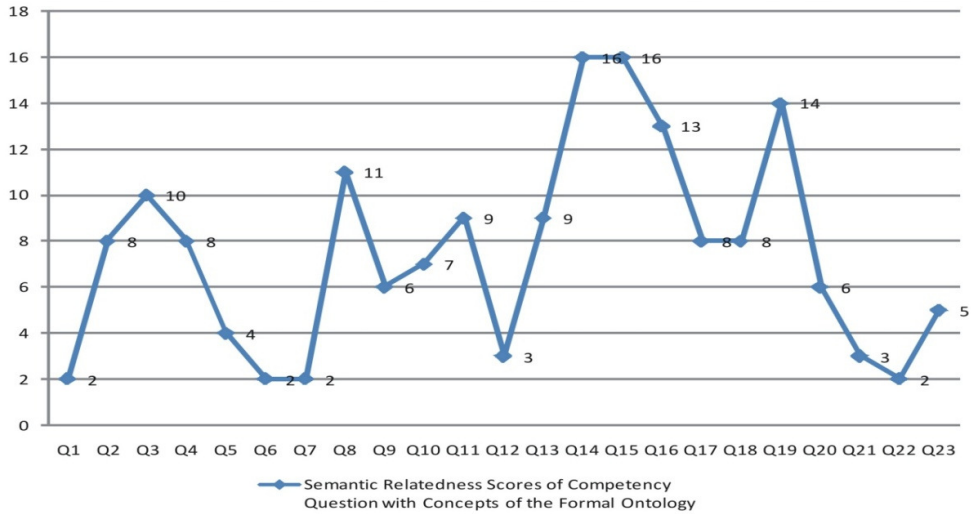

Figure 5: Chart of the mapping of competency questions with concepts of the formal ontology

Table 7: Comparative table of this study with related research

| Authors | Ontology Building Methodology | Framework / Algorithm | Level of Specification | Implementation Details | | |
|---|---|---|---|---|---|---|
| | | | | Languages | Platform | Open / Proprietary |
| Sanati and Lu [6] | unreported | none | conceptual | none | none | |
| Apostolou et al. [1], [2] | unreported | none | conceptual | unreported | ONTOGOV | proprietary |
| Xiao et al. [3] | unreported | none | conceptual, formal | OWL | unreported | |
| Chen et al. [8] | unreported | none | conceptual | none | none | |
| (Our Work) | Uschold and King | adaptive framework | conceptual, semi-formal, formal | OWL | Protégé, Jcreator, JGrasp | open-source |
| Gugliotta et al. [9] | derived from DOLCE upper level ontology | none | conceptual, semi-formal | OCML | unreported | |
| Sabucedo and Rifon [10] | Methontology | none | semi-formal | none | none | |
| Sabucedo et al. [7] | unreported | none | semi-formal, formal | OWL | unreported | |
| Puustjarvi [11] | unreported | none | semi-formal | none | none | |

In Table 7 and Fig. 6, it appears that only 3 (33%) researches out of 9, including this study have reported the ontology building methodology they have used to build the e-government domain specific ontology models. Furthermore, although Sabucedo and Rifon [10] and Gugliotta et al. [9] have disclosed the ontology building techniques they have used in their studies, little information is provided in these studies to explain how the methodologies can be applied to build the proposed ontology models for a complex public administration system. Similarly, amongst the 9 studies reported in Table 7, only 4 (44%) have reported the formal specification of the proposed e-government domain specific ontology models and programming codes in Semantic Web ontology languages. Table 6 shows that only this study has provided complete information for all the comparison criteria; most of the related works studied have reported their proposed e-government domain specific ontology models at the conceptual level only (78% in Fig. 6), with little information provided on the ontology building methodologies employed and the implementation details. In Table 7, it is also shown that only this study has used a step-by-step framework adopted from an existing ontology building methodology to develop government domain ontology. Further information on the comparison criteria and results is depicted in Fig. 6.

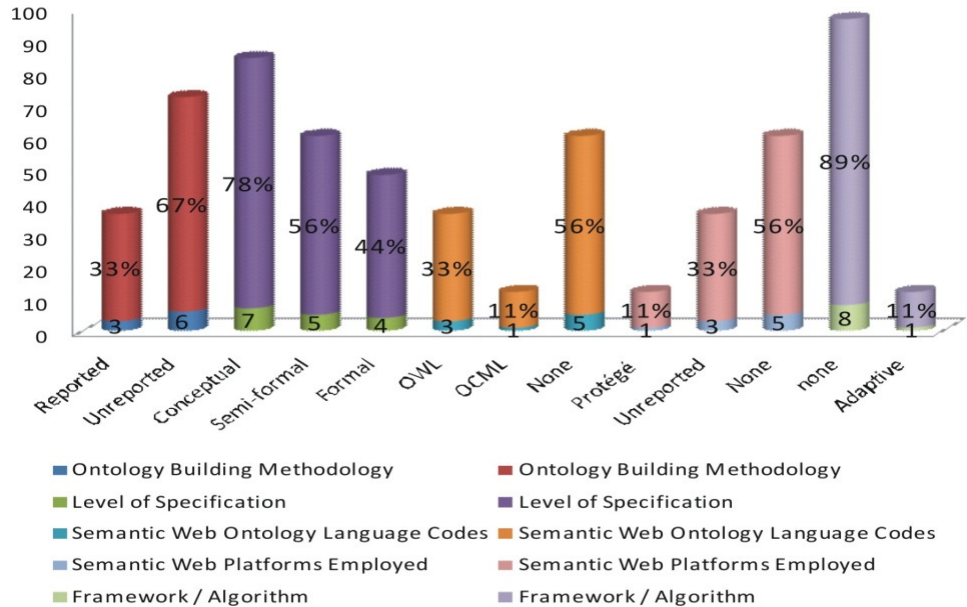

Figure 6: Chart of the comparison of this study with related research

## 4. CONCLUSION

This research has presented a detailed application of the Uschold and King [12] ontology building methodology for the development of semantic ontology models including conceptual/domain, semi-formal and formal ontologies in a government service domain. The research represents a practical case study application of an existing ontology building methodology for developing semantic ontology models in e-government. This may promote the use of existing ontology building methodologies in the Semantic Web development processes of government domain ontologies, facilitate the repeatability of the resulting domain ontologies in other e-government researches and projects, and strengthen the adoption of semantic technologies in e-government.

Although the study has focused on developing semantic ontology models for e-government applications, it also represents a contribution in the ontology engineering field in general and the Semantic Web domain in particular. In fact, the framework and techniques employed provides a practical application process of the Uschold and King [12] methodology which might be easily repeated in other domains of knowledge to build domain ontologies.

## Authors

[1]**J.V. Fonou Dombeu** is a PhD candidate at the School of Computer, Statistical and Mathematical Sciences at the North-West University, South Africa and a Lecturer in the Department of Software Studies at the Vaal University of Technology, South Africa. He received an MSc. in Computer Science at the University of KwaZulu-Natal, South Africa, in 2008, BSc. Honour's and BSc. in Computer Science at the University of Yaoundé I, Cameroon, in 2002 and 2000 respectively. His research interests include: Biometric for Personal Identification, Ontology, Agent Modelling, and Semantic Knowledge representation in e-Government. Mr. Fonou-Dombeu has published articles on Semantic Web development in e-Government and presented papers at international conferences in South Africa, France and Slovenia. 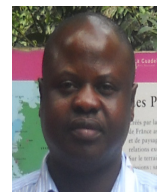

[2]**M. Huisman** is a professor of Computer Science and Information Systems at the North-West University (Potchefstroom Campus) where she teaches software engineering, management information systems, and decision support systems. She received her Ph.D degree in Computer Science and Information Systems at the Potchefstroom University for CHE in 2001. Magda is actively involved in research projects regarding systems development methodologies. Her research has appeared in journals such as MISQ, Information & Management, International Journal on Computer Science and Information Systems, and Lecture Notes in Computer Science. She has presented papers at international conferences in China, Australia, Switzerland, Canada, Japan, Latvia and Slovenia. Her current research interests are in systems development methodologies and the diffusion of information technologies. 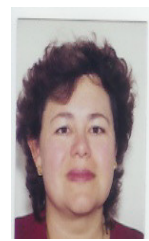